\documentclass[letterpaper, 10 pt, conference]{ieeeconf}  

\IEEEoverridecommandlockouts                              

\usepackage[T1]{fontenc}
\usepackage{mathtools}
\usepackage{amssymb}  
\usepackage{graphicx}
\usepackage{subcaption}
\usepackage{hyperref}

\usepackage[backend=biber,sorting=none,style=numeric, firstinits]{biblatex}
\title{\LARGE \bf
Very High Frequency Interpolation for Direct Torque Control
}

\author{Rafael Kourdis$^{1, 2}$, Maciej Stępień$^{2, 3}$, Jérôme Manhes$^{2}$, \\ Nicolas Mansard$^{2, 3}$, Steve Tonneau$^{1}$, Philippe Souères$^{2}$ and Thomas Flayols$^{2}$
\thanks{$^{1}$University of Edinburgh, School of Informatics, UK}%
\thanks{$^{2}$LAAS-CNRS, Université de Toulouse, France}%
\thanks{$^{3}$Artificial and Natural Intelligence Toulouse Institute, France}
}

\begin{document}

\maketitle
\thispagestyle{empty}
\pagestyle{empty}

\begin{abstract}
Torque control enables agile and robust robot motion, but deployment is often hindered by instability and hardware limits. Here, we present a novel solution to execute whole-body linear feedback at up to 40 kHz on open-source hardware.  We use this to interpolate non-linear schemes during real-world execution, such as inverse dynamics and learned torque policies. Our results show that by stabilizing torque controllers, high-frequency linear feedback could be an effective route towards unlocking the potential of torque-controlled robotics. 
\end{abstract}

\section{Introduction}

Performing robot control directly in torques could unlock better performance than relying on low-level joint controllers such as PD feedback  \cite{evocontrol} \cite{loco_torques}. However, deploying torque controllers on hardware is often difficult and requires high frequency feedback to achieve stability \cite{pierrealex} \cite{mpc_approx}.

In our work, we introduce a method to reliably execute non-linear torque controllers on our open-source biped, Bolt. To achieve this, we implement a robot-agnostic solution to perform full-body linear feedback at up to 40 kHz, an order of magnitude faster than existing techniques \cite{mpc_approx}. We utilize linear feedback to interpolate torque controllers between evaluations, and execute the interpolation at a high frequency.

We perform experiments with controllers based on reinforcement learning and classical inverse dynamics. Our results show that this approach is an effective method to deploy controllers that are otherwise unstable, without ad-hoc joint impedance to dampen robot dynamics. Our work paves the way for robotics controlled directly in torques, raising the bar in terms of achievable performance, reactivity and robustness \cite{evocontrol}.

\begin{figure}
    \centering
    \begin{subfigure}[T]{.245\textwidth}
        \hspace{0.2cm}
        \includegraphics[width=.9\linewidth]{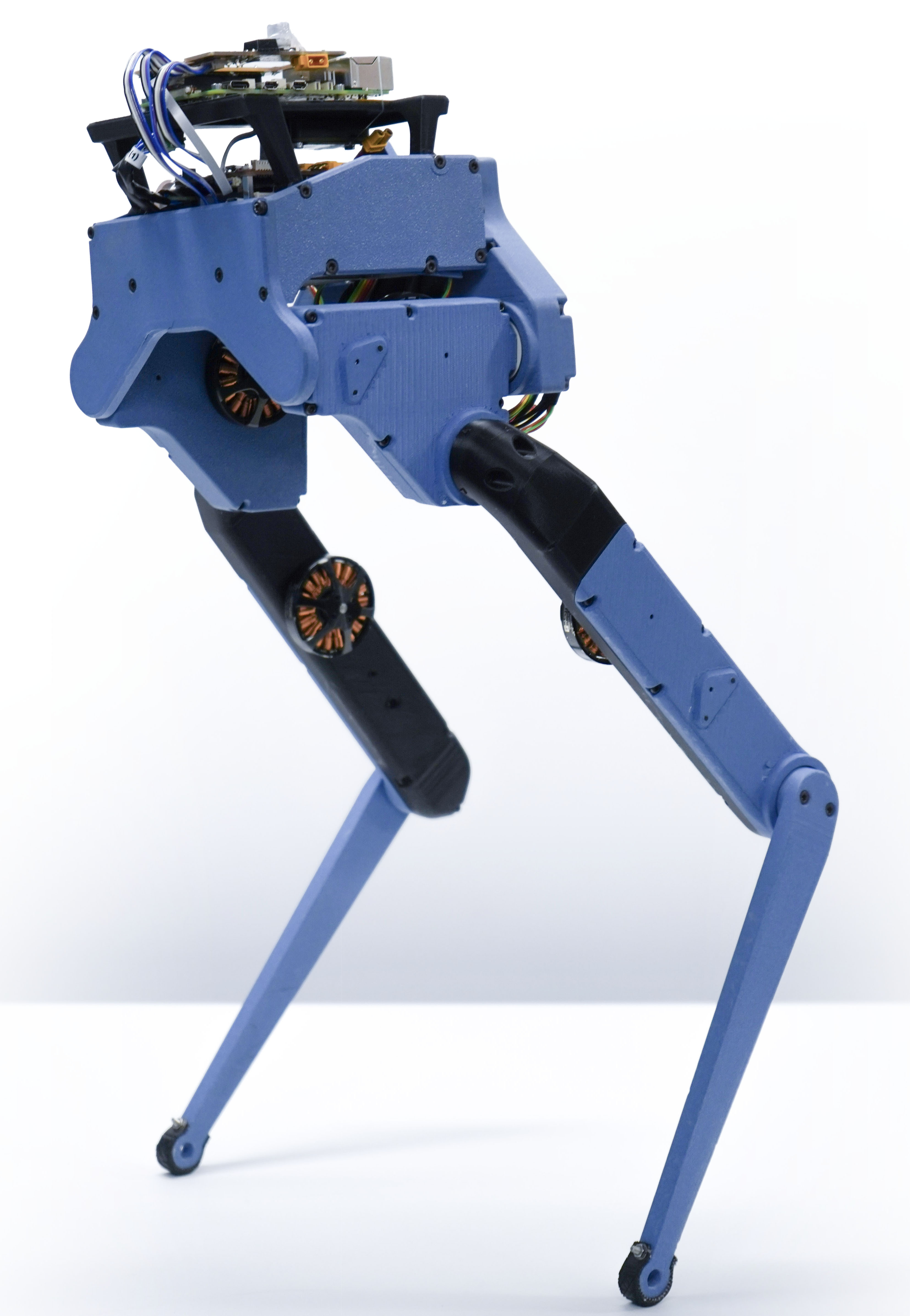}
    \end{subfigure}
    \hspace{0.3cm}
    \begin{subfigure}[T]{.15\textwidth}
        \includegraphics[width=.9\linewidth]{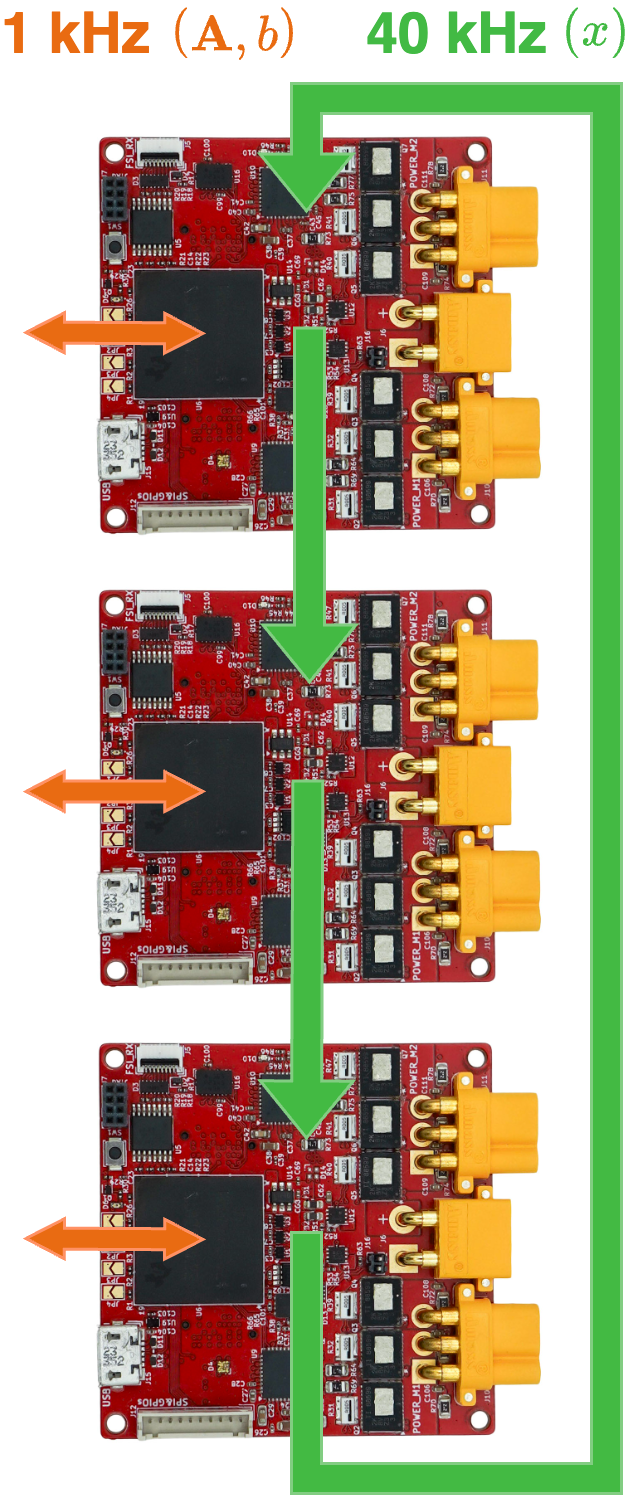}
    \end{subfigure}
    \caption{Our open-source Bolt biped (left). Diagram of our control scheme running on \textit{uOmodri}: 40 kHz full-body state feedback with linear controller updating at 1 kHz (right). }
    \label{fig:bolt_and_diag}
\end{figure}

\section{Method}
Suppose a non-linear, time- and state-dependent torque controller for a robot with $n$ joints: $\tau = \tau(x;\,t)$, where $x\in\mathbb{R}^{2n}$ is a state vector containing joint angles and velocities.

Between controller evaluations, we can obtain a linear interpolation of $\tau$ as $x$ evolves, via first-order approximation:
\begin{equation}
    \tau(x_k+\Delta x;\, t_k)\approx \tau(x_k;\,t_k) + \left.\frac{\partial\tau}{\partial x}\right|_{x_k} \Delta x
\end{equation}
where $t_k$ is the controller evaluation time. This can be written as: $\tau = \textbf{A}x+b$, with $\textbf{A}$ a $n\times2n$ matrix and $b\in\mathbb{R}^n$.

To achieve stability of a torque controller on hardware, a linear interpolation can be used to perform higher frequency feedback. At each timestep, a new linearization is obtained by differentiating the non-linear controller. The resulting $\textbf{A}$ and $b$ are then sent to the robot.

This idea is explored in \cite{mpc_approx}, where 2 kHz linear feedback is used to deploy a slow MPC controller on a humanoid. However, we show that linear interpolation is useful in general, and not limited to controllers with Riccati gains.

\section{Implementation}
In order to interpolate controllers at a high frequency, we introduce a hardware implementation of full-body linear feedback at up to 40 kHz. We base our work on the \textit{uOmodri} board, a high-performance motor driver designed for academic research by the Open Dynamic Robot Initiative \cite{odri}. The \textit{uOmodri} can perform torque control at 40 kHz for two brushless motors.

To achieve full-body feedback on a robot with more than two joints, we implement a novel state-sharing scheme via a board-to-board daisy chain network (Fig. \ref{fig:bolt_and_diag}, right). This is necessary due to the off-diagonal terms in $\textbf{A}$. Effectively, all motor drivers must know the full robot state at each cycle in order to compute a torque.

In our implementation, we support a maximum feedback frequency of 40 kHz, which can be decimated at the firmware level. $\textbf{A}$ and $b$ can be updated by the computer at up to \mbox{1 kHz}. Both our hardware and control scheme are open source.

\begin{figure}[h]
    \centering
    \hspace*{-1.6mm}\includegraphics[width=0.5\textwidth]{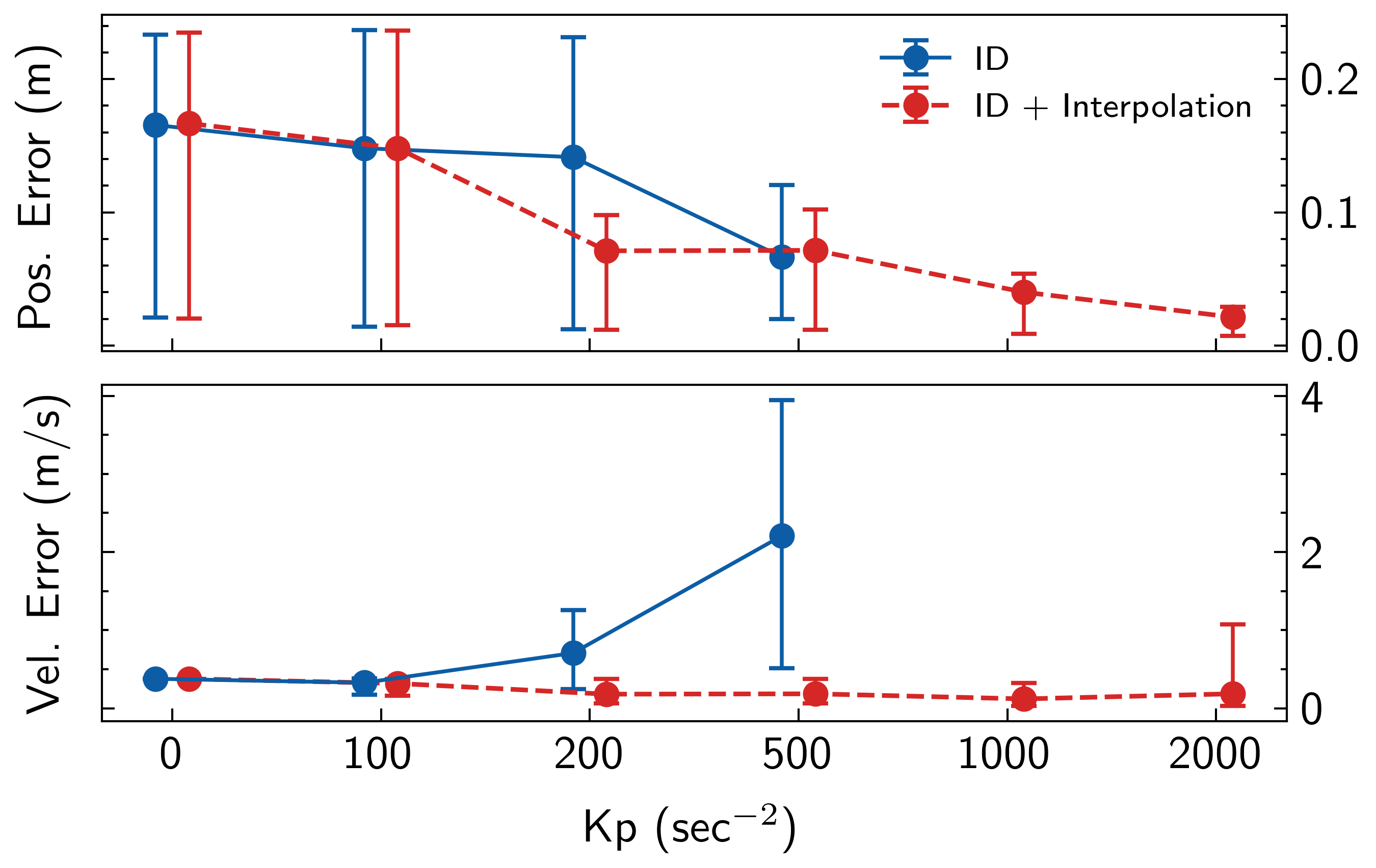}%
    \caption{Tracking errors for different $K_p$ of the the ID controller. Error bars indicate the 5th and 9th percentiles.}%
    \label{id_error_distr}%
\end{figure}

\section{Experiments}
To investigate the effectiveness of linear controller interpolation, we evaluate multiple controllers on a trajectory tracking task. We use a modified version of the open-source 6-DoF biped robot, Bolt \cite{odri} (Fig. \ref{fig:bolt_and_diag}, left), with three \textit{uOmodri} boards and a Raspberry Pi 5 for controller computations.

We define the trajectory reference so that Bolt's right foot tracks a circle while the base is fixed. Our aim is to explore whether using high-frequency interpolation improves tracking stability or performance.

\subsection{Inverse Dynamics (ID)}

Given the desired foot trajectory: $p^* = p^{foot}(t)\in\mathbb{R}^3$, we design an instantaneous tracking controller of the form:
\begin{equation}
    \tau = ID(q, v, a^{*})
    \label{id_eqn}
\end{equation}
where $q, v \in \mathbb{R}^n$ are joint position and velocity vectors, $ID$ the inverse dynamics function and $a^{*}\in\mathbb{R}^n$ a joint-space acceleration such that the foot tracks $p^*(t)$:
\begin{equation}
    \ddot{p}=\dot{\mathbf{J}}v + \mathbf{J}a^* =  \ddot{p}^{*} + K_p(p^{*} - p) + K_d(\dot{p}^* - \dot{p})
\end{equation}
with $\mathbf{J}=\mathbf{J}(q)$ the foot kinematic Jacobian. The $K_p$ and $K_d$ diagonal terms act as task-space feedback. We set $K_d = \sqrt{2K_p}$ so the foot error dynamics are critically damped. This type of controller is noted in \cite{pierrealex} as difficult to deploy on a quadruped with characteristics similar to Bolt. This is due to the fast dynamics of the low-inertia, quasi-direct-drive actuators and lightweight legs.

We evaluate the controller at 500 Hz with different $K_p$ gains, trading off tracking stiffness with stability. For the interpolation experiments, we additionally linearize (\ref{id_eqn}) around $q$ and $v$ to obtain $\textbf{A}$ and $b$ at 500 Hz. These are used to perform linear feedback at 40 kHz on the hardware.

\begin{figure}[h]
    \centering
    \includegraphics[width=0.48\textwidth]{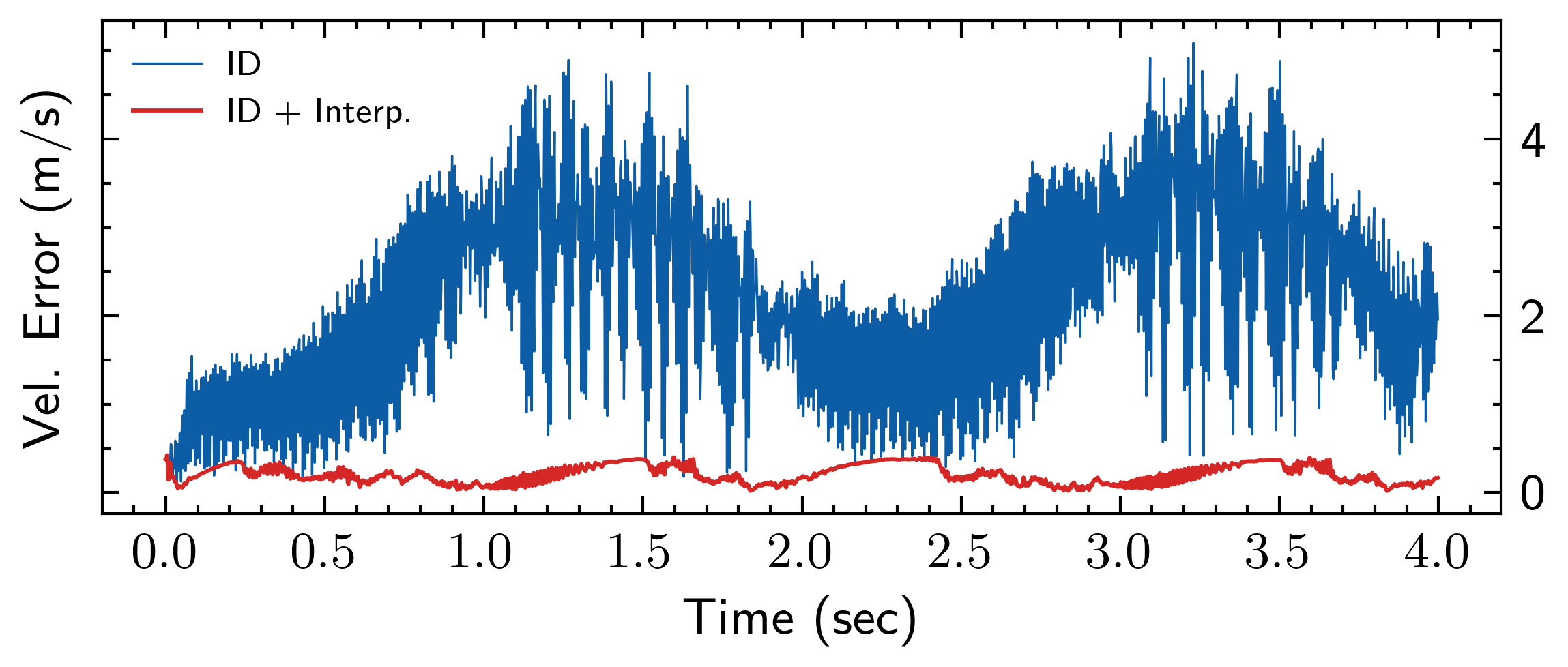}%
    \caption{Velocity tracking error over time (ID controller with $K_p=500\,s^{-2}$).} %
    \label{vel_err_time}%
\end{figure}

\begin{figure}[h]
    \centering
    \includegraphics[width=0.48\textwidth]{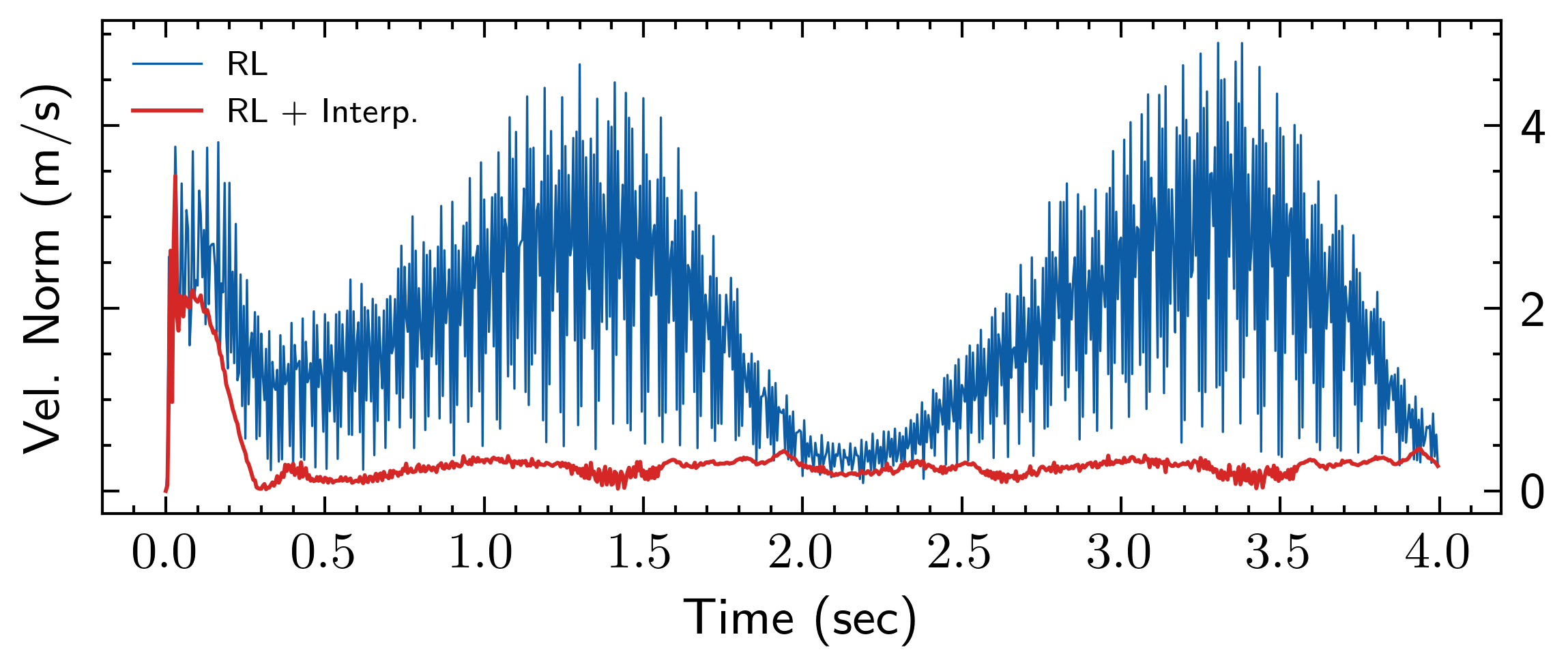}%
    \caption{Norm of foot velocity over time (RL controller). The initial jump is due to the foot starting position being far from the commanded trajectory.} %
    \label{vel_nnet}%
\end{figure}

As shown in Fig. \ref{id_error_distr}, the position error during tracking improves when $K_p$ is increased (Fig. \ref{id_error_distr}). However, large values result in violent vibrations. As a result, we choose not to deploy controllers with $K_p$ above 500\,sec$^{-2}$ to prevent hardware damage. On the other hand, interpolation stabilizes the system, as reflected by the smaller range of velocity errors (Fig. \ref{id_error_distr}). This allows us to safely test higher $K_p$ values, achieving significantly better tracking accuracy. The time domain plot (Fig. \ref{vel_err_time}) confirms that lack of interpolation results in high frequency shaking, rendering the controller unsuitable for deployment on the robot.

\subsection{Reinforcement Learning (RL)}

We also investigate the case where a learned neural network policy is used to generate torque commands: $ \pi_\theta(\tau \,|\,x, p^*(t))$, with $p^*(t)$ the desired foot trajectory. We use PPO \cite{ppo} to optimize $\theta$ for the task of end-effector position tracking in Isaac Lab \cite{isaaclab}. The policy runs at 200 Hz, similar to the work in  \cite{loco_torques}.

In the baseline case, the network output is the torque command for the robot. When linear feedback is used between network evaluations, we obtain $\textbf{A}$ and $b$ by computing the Jacobian of the output with respect to the robot state. We run feedback on the hardware at 40 kHz. Between network evaluations, the tracking command is held constant.

Without interpolation, the RL controller performs poorly, similar to the high $K_p$ ID controllers. High-frequency vibration (Fig. \ref{vel_nnet}) renders it unsuitable for hardware deployment. However, linear feedback stabilizes the controller, enabling effective sim-to-real transfer. 

\section{Future Work}

In this work, we perform preliminary experiments to validate our open-source implementation of high-frequency linear feedback. Open questions include the impact of lowering the feedback rate or changing parameters such as low-level velocity filters, and the applicability of our method to tasks with discontinuous dynamics, such as locomotion. We would also like to extend our method to include base attitude from inertial sensors in the feedback state.

\addtolength{\textheight}{-12cm}   

\printbibliography

\end{document}